\documentclass[10pt,twocolumn,letterpaper]{article}

\usepackage{iccv}
\usepackage{times}
\usepackage{epsfig}
\usepackage{graphicx}
\usepackage{amsmath}
\usepackage{amssymb}
\usepackage{caption}
\usepackage{subcaption}

\usepackage[breaklinks=true,bookmarks=false]{hyperref}

\usepackage{amsmath,amsfonts,bm}

\def\eqref#1{equation~\ref{#1}}

\def\1{\bm{1}}

\DeclareMathAlphabet{\mathsfit}{\encodingdefault}{\sfdefault}{m}{sl}
\SetMathAlphabet{\mathsfit}{bold}{\encodingdefault}{\sfdefault}{bx}{n}

\usepackage{url}

\usepackage{graphicx}
\graphicspath{{./figs/}}

\usepackage{soul}

\newcommand\cifar{\texttt{CIFAR10{ }}}
\newcommand\cifarh{\texttt{CIFAR10H{ }}}

\newcommand\cifarhn{\texttt{CIFAR10H}}

\newcommand\ten{\texttt{c10{ }}}
\newcommand\vfour{\texttt{v4{ }}}
\newcommand\vsix{\texttt{v6{ }}}
\newcommand\tenh{\texttt{c10H{ }}}

\newcommand\smallsec[1]{\paragraph{#1.}}

\newcommand\blfootnote[1]{%
  \begingroup
  \renewcommand\thefootnote{}\footnote{#1}%
  \addtocounter{footnote}{-1}%
  \endgroup
}

\iccvfinalcopy

\def\iccvPaperID{6284} 

\ificcvfinal\fi

\begin{document}

\title{Human uncertainty makes classification more robust}

\author{Joshua C. Peterson*, Ruairidh M. Battleday*, Thomas L. Griffiths, Olga Russakovsky\\
Princeton University, Department of Computer Science\\
{\tt\small \{joshuacp,battleday,tomg,olgarus\}@cs.princeton.edu}
}

\maketitle
\thispagestyle{empty}

\begin{abstract}
The classification performance of deep neural networks has begun to asymptote at near-perfect levels. However, their ability to generalize outside the training set and their robustness to adversarial attacks have not. In this paper, we make progress on this problem by training with full label distributions that reflect human perceptual uncertainty. We first present a new benchmark dataset which we call \cifarhn, containing a full distribution of human labels for each image of the \texttt{CIFAR10} test set. We then show that, while contemporary classifiers fail to exhibit human-like uncertainty on their own, explicit training on our dataset closes this gap, supports improved generalization to increasingly out-of-training-distribution test datasets, and confers robustness to adversarial attacks. 
\end{abstract}

\section{Introduction}
On natural-image classification benchmarks, state-of-the-art convolutional neural network (CNN) models have been said to equal or even surpass human performance, as measured in terms of ``top-1 accuracy''---the correspondence between the most probable label indicated by the model and the ``ground truth'' label for a test set of held-out images. As accuracy gains have begun to asymptote at near-perfect levels \cite{shake-shake}, there has been increasing focus on out-of-training-set performance---in particular, the ability to generalize to related stimuli \cite{recht2018cifar}, and robustness to adversarial examples \cite{kurakin2016adversarial}. On these tasks, by contrast, CNNs tend to perform rather poorly, whereas humans continue to perform well.\blfootnote{* contributed equally}

\begin{figure}[!t]
    \centering
    \includegraphics[width=0.92\linewidth]{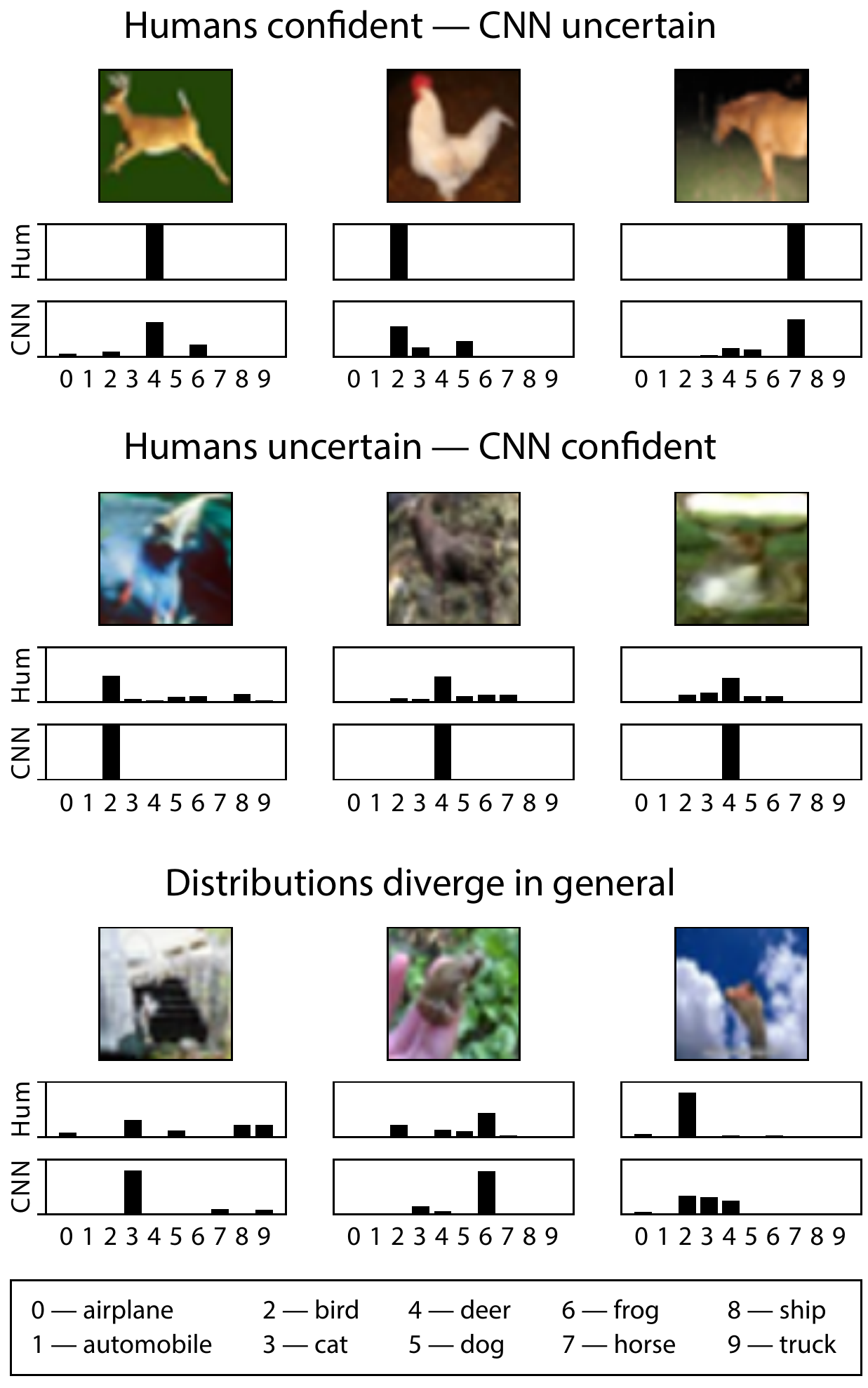}
    \caption{\texttt{CIFAR10} images for which humans and our best traditionally-trained CNN (Shake-Shake~\cite{shake-shake}) agree in their top guess, but systematically differ over other choices.}
    \label{fig:example_stim}
\end{figure}

To redress this problem, and provide a better standard for training classifiers, we suggest an alternative objective: not just trying to capture the most likely label, but trying to capture the full distribution over labels. Errors in classification can be just as informative as the correct answers---a network that confuses a dog with a cat, for example, might be judged to generalize better than one that confuses it with a truck (see \cite{awad2018moral}). Indeed, consider the examples shown in Figure \ref{fig:example_stim}, in which the CNN can be  underconfident, overconfident, or systematically incorrect, and yet receive a perfect accuracy score. Capturing this similarity structure is a key part of effective generalization \cite{Hinton2015DistillingTK}, and an important consideration when building classification models for real-world applications, for example, object avoidance in driverless cars.

Predicting more complete distributions of labels requires first measuring those distributions. Given that we cannot directly extract ground truth perceptual similarity from the world, human categorization behavior is a natural candidate for such a comparison. Indeed, there is often a lack of human consensus on the category of an object, and human errors often convey important information about the structure of the visual world \cite{lakoff2008women}. Beyond complementing training paradigms, collecting these full label distributions from humans to better model human biases and predict their errors is interesting in itself---this time, for example, to help a driverless car infer the actions of nearby human drivers. Finally, although there has been much work scaling the number of images in datasets \cite{scaling_Hestness}, and investigating label noise \cite{RolnickVBS17_label, ghosh2017robust_label, NIPS2017_Vahdat_label}, little effort has been put into identifying the benefits from increasing the richness of (informative) label distributions for image classification tasks. 

To these ends, we make the following contributions:
\begin{itemize}
    \item We present a novel soft-label dataset which we call \cifarhn, comprising full label distributions for the entire $10{,}000$-image \texttt{CIFAR10} test set, utilizing over $500$k crowdsourced human categorization judgments.
    \item We show that when state-of-the-art CNN classifiers are trained using these soft labels, they generalize better to out-of-sample datasets than hard-label controls. 
    \item We present a performance benchmark assessing model fit to human labels, and show that models trained using alternative label distributions do not approximate human uncertainty as well. 
    \item We show that when CNNs are trained to perform well on this benchmark they are significantly more resistant to adversarial attacks.
\end{itemize}
Taken together, our results support more fine-grained evaluations of model generalization behavior and demonstrate the potential utility of one method for integrating human perceptual similarity into paradigms for training classifiers.

\section{Related Work}
\label{section:related_work}
\smallsec{Hierarchical Classification} Work on using class confusion or hierarchy to improve classification accuracy or robustness dates back to early works of \textit{e.g.,} Griffin and Perona~\cite{griffin2008learning}, Marszalek and Schmid~\cite{marszalek2007semantic}, or Zweig and Weinshall~\cite{zweig2007exploiting}. Class label hierarchies have been used to enable \textit{e.g.,} sharing of representations~\cite{torralba2007sharing,fergus2010sharing,hwang2011sharing}, effective combination of models~\cite{jia2013hierarchies}, or improved accuracy of classification through hierarchical prediction~\cite{lampert2011multi,deng2014relation}. Benchmarks have occasionally proposed using hierarchical metrics for evaluation (\textit{e.g.,} the hierarchical error rate of ILSVRC 2010 and 2011~\cite{russakovsky2015imagenet}). Overall though the dominant paradigm has focused on evaluating the top-K accuracy rather analyzing the errors of the system, and the hierarchical structure has been used mostly for training. We argue it is time to rethink this. First, modern large-scale open-world complex datasets no longer guarantee non-overlapping object classes~\cite{krishna2007visualgenome}, making hierarchical class confusion particularly meaningful. Second, existing methods are becoming remarkably good at top-K accuracy, so an increasing focus on their robustness with regard to adversarial examples~\cite{szegedy2013intriguing,goodfellow2014explaining,biggio2018wild} or distributional shift \cite{torralba2011unbiased,recht2018cifar} is warranted. In this work we present to our knowledge the first large-scale evaluation of generalization to human uncertainty in image classification.

\smallsec{Knowledge Distillation} The label hierarchies used to aid recognition can be manually constructed~\cite{chilton2013cascade,bragg2013taxonomy}, derived from linguistic knowledge bases~\cite{frome2013devise,fergus2010sharing}, or learned automatically~\cite{griffin2008learning,Hinton2015DistillingTK}. Our work is closest to the former (manual construction), although instead of explicitly constructing a class hierarchy we rely on human confusion between the classes to infer the relationship between the classes for a given image. While being derived from \emph{human} confusion, our work bears some resemblance to the knowledge distillation approach of~\cite{Hinton2015DistillingTK}. In knowledge distillation, these labels are provided by the smoothed softmax probabilities from a pre-trained classification model. When soft labels are combined with ground truths, a form of model transfer and compression is achieved, because the softmax probabilities carry crucial information. The rationale for this process is similar to our own: networks (and humans) gain great robustness from distilling important information about similarity structure into the distributions we infer over images and their categories. However, the use of a network to provide them (\textit{i.e.,} the standard application of knowledge distillation) is itself problematic without a gold standard to compare to: there is no guarantee that the similarity structure a model has learned is correct.

\smallsec{Soft Labels} One of the core contributions of our work is around using the soft labels provided through human confusion as a replacement for one-hot label encodings. Several methods have been proposed as alternatives to one-hot encodings, \textit{e.g.,} using heuristics to smooth the top-1 label during large-scale 1000+ way classification~\cite{szegedy2016rethinking} or incorporating test-time human uncertainty into a collaborative computer vision system~\cite{branson2010visual}. \textit{mixup} \cite{zhang2017mixup} is another recently developed method for automatically generating soft labels based on convex combinations of pairs of examples and their hard labels, and has been shown to improve generalization and adversarial robustness while reducing memorization. However, since the linearity constraint is constant across all pairs of classes, and the labels are one-hot, it is difficult to see how the softness in such labels is a full measure of perceptual likeness.

\smallsec{Human studies} Lastly, there are a number of studies that also use human experts to provide distributional information over training labels in related classification fields, such as medical diagnosis systems \cite{nguyen2015combining, nguyen2014learning}. While the theoretical cases these studies present support our own, they do not provide a large-scale testbed for evaluation of other classification models. Notably, the human uncertainty labels frequently don't need to be explicitly collected but will become automatically available in the process of data collection. Much of crowdsourcing work focuses on \emph{reconciling} human labels and mitigating their disagreement (c.f., Kovashka et al.~\cite{kovashkaFT16} for a survey). Our approach proposes utilizing these human disagreements to improve the accuracy and robustness of a model, complementing existing work aimed at leveraging ``errors'' in human labeling~\cite{krishna2016embracing}.

\section{From Labels to Label Distributions}
\label{section:theory}

The standard practice for image classification tasks is to train using ``ground truth" labels provided in common benchmark datasets, for example, \texttt{ILSVRC12} \cite{russakovsky2015imagenet}, and \texttt{CIFAR10} \cite{krizhevsky2009learning}, where the ``true" category for each image is decided through human consensus (the {\em modal} choice) or by the database creators. Although a useful simplification in many cases, we suggest that this approximation introduces a bias into the learning framework that has important distributional implications. To see this, first consider the standard loss minimization objective during training given below:
\begin{equation}
\min_{\theta} \sum_{i=1}^{n} \mathcal{L}(f_{\theta}, x_i, y_i),
\end{equation}
in which the loss $\mathcal{L}$ for a model with parameters $\theta$ is minimized with respect to observed data samples $\{x_i, y_i\}_{i=1}^n$. Our goal in training a model in this way is to generalize well to unseen data: to minimize the expected loss over unobserved labels given observed images $\{x_j\}_{j=1}^m$ drawn from the same underlying data distribution in the future:
\begin{equation}
\frac{1}{m} \sum_{j=1}^m \sum_c \mathcal{L}(f_{\theta}, x_j, y_j = c) \, p(y_j = c|x_j).
\end{equation}
When we consider the second term in this product, we can see that using modal labels during dataset construction would only be an optimal estimator if for any stimulus $x$, the underlying conditional data distribution $p(y|x)$ is zero for every category $c$ apart from the one assigned by human consensus. By contrast, when we consider the network and human confusions seen in Figure \ref{fig:example_stim}, we can see there do exist cases in which this assumption violates human allocation of probabilities. 

How, then, can we reach a more natural approximation of $p(y|x)$? For some problems, it is easy to just sample from some real set of data $p(x,y)$, but for image classification, we must rely on humans as a gold standard for providing a good estimate of $p(y|x)$. If we expect the human image label distribution $p_{\rm hum}(y|x)$ to better reflect the natural distribution over categories given an image, we can use it as an improved estimator for $p(y|x)$. 

In the case where $f_{\theta}(x)$ is a distribution $p_{\theta}(y|x)$ and $\mathcal{L}(f,x,y)$ is the negative log-likelihood, the expected loss reduces to the cross-entropy between the human distribution and that predicted by the classifier:
\begin{equation}
    -\frac{1}{m}\sum_{j=1}^m\sum_c p_{\rm hum}(y_j = c|x_j) \log{ p_{\theta}(y_j=c|x_j)}.
\end{equation}
This implies that the optimal strategy for gathering training pairs $\{x_i, y_i\}_{i=1}^n$ is to sample them from $p_{\rm hum}(y|x)$. Our dataset provides this distribution directly, so that models may be trained on human labels or evaluated against them, or better approximations of $p(y|x)$ for natural images be found. In turn, better approximation of this underlying data distribution should be expected to give better generalization and robustness.

\section{Dataset Construction}
While larger-scale popular datasets such as ImageNet~\cite{russakovsky2015imagenet}, Places~\cite{zhou2017places}, or COCO~\cite{lin_microsoft_2014} might seem like the best starting point, \texttt{CIFAR10} in particular has several unique and attractive properties. First, the dataset is still of enough interest to the community that  state-of-the-art image classifiers are being developed on it \cite{shake-shake,huang2018gpipe}. Second, the dataset is small enough to allow us to collect \emph{substantial} human data for the entire test set of images. Third, the low resolution of the images is useful for producing variation in human responses. Human error rates for high resolution images with non-overlapping object categories are sufficiently low that it is hard to get a meaningful signal from a relatively small number of responses. Finally, \texttt{CIFAR10} contains a number of examples that are close to the category boundaries, in contrast with other datasets that are more carefully curated such that each image is selected to be a good example of the category. Our final \texttt{CIFAR10H} behavioral dataset consists of $511{,}400$ human categorization decisions over the $10{,}000$-image testing subset of \texttt{CIFAR10} (approx. $50$ judgments per image).

\subsection{Image Stimuli}
We collected human judgments for all $10{,}000$ $32\!\times\!32$ color images in the {\it testing} subset of \texttt{CIFAR10}. This contains $1{,}000$ images for each of the following 10 categories: \textit{airplane}, \textit{automobile}, \textit{bird}, \textit{cat}, \textit{deer}, \textit{dog}, \textit{frog}, \textit{horse}, \textit{ship}, and \textit{truck}. This allows us to evalulate models pretrained on the \texttt{CIFAR10} training set using the same testing images, but in terms of a different distribution over labels, detailed in the next section.

\subsection{Human Judgments}
We collected $511{,}400$ human classifications over our stimulus set via Amazon Mechanical Turk \cite{buhrmester2011amazon}---to our knowledge, the largest of its kind reported in a single study to date. In the task, participants were asked to categorize each image by clicking one of the 10 labels surrounding it as quickly and accurately as possible (but with no time limit). Label positions were shuffled between candidates. After an initial training phase, each participant ($2{,}571$ total) categorized $200$ images, $20$ from each category. Every $20$ trials, an obvious image was presented as an attention check, and participants who scored below $75\%$ on these were removed from the final analysis ($14$ total). We collected $51$ judgments per image on average (range: $47 - 63$). Average completion time was $15$ minutes, and workers were paid $\$1.50$ total. Examples of distributions over categorization judgments for a selection of images is shown in Figure \ref{fig:example_stim}.

\section{Generalization Under Distributional Shift}
\label{section:primary_training}
Our general strategy is to train a range of classifiers using our soft labels and assess their performance on held-out validation sets and a number of generalization datasets with increasing distributional shift. We expect the human information about image label uncertainty to be most useful when test datasets are increasingly out-of-distribution.

\subsection{Setup}
\label{model-steup}

\smallsec{Models} We trained eight CNN architectures (VGG~\cite{simonyan2014very}, ResNet~\cite{he2016deep}, Wide ResNet~\cite{zagoruyko2016wide}, ResNet preact~\cite{he2016identity}, ResNext~\cite{xie2016aggregated}, DenseNet~\cite{huang2017densely}, PyramidNet~\cite{han2017deep}, and Shake-Shake~\cite{shake-shake}) to minimize the crossentropy loss between softmax outputs and the full human-label distributions for images in \texttt{CIFAR10H}. The models were trained using \texttt{PyTorch} \cite{paszke2017automatic}, adapting the repository found in the footnote.\footnote{\url{github.com/hysts/pytorch_image_classification}; \\
model identifiers \texttt{vgg\_15\_BN\_64}, \texttt{resnet\_basic\_110}, \texttt{wrn\_28\_10},\\
\texttt{resnet\_preact\_bottleneck\_164},  \texttt{resnext\_29\_8x64d},\\ \texttt{densenet\_BC\_100\_12}, \texttt{pyramidnet\_basic\_110\_270}, \\ \texttt{shake\_shake\_26\_2x64d\_SSI\_cutout16} (output folder names).} For each architecture, we train 10 models using 10-fold cross validation (using $9{,}000$ images for training each time) and at test time average the results across the 10 runs. We use $k$-fold instead of a single validation set in order to obtain more stable results. We used the default hyperparameters in the repository for all models, following \cite{recht2018cifar} for the sake of reproducibility, except for the learning rate. We trained each model for a maximum of 150 epochs using the Adam \cite{adam} optimizer, and performed a grid-search over base learning rates $0.2$, $0.1$, $0.01$, and $0.001$ (we found $0.1$ to be optimal in all cases).

\smallsec{Test Datasets}
A key prediction from section \ref{section:theory} is that the uncertainty in our labels will be increasingly informative when generalizing to increasingly out-of-training-sample distributions. We test this prediction empirically by examining generalization ability to the following datasets:

{\bf \texttt{CIFAR10}:} This is the standard within-dataset evaluation. Since our \texttt{CIFAR10H} soft labels are for the \texttt{CIFAR10} test set, here we use the 50{,}000-images of the standard \texttt{CIFAR10} training set to instead evaluate the models.

{\bf \texttt{CIFAR10.1v6,v4}:} These are two 2{,}000-image near-sample datasets created by \cite{recht2018cifar} to assess overfitting to \texttt{CIFAR10} ``test'' data often used for validation. The images are taken from TinyImages \cite{torralba200880}  and match the sub-class distributions in \texttt{CIFAR10}. \texttt{v6} has 200 images per class while \texttt{v4} is the original class-unbalanced version (90\% overlap).

{\bf \texttt{CINIC10}:} This is an out-of-sample generalization test. The \texttt{CINIC10} dataset collected by \cite{cinic} contains both \texttt{CIFAR10} images and rescaled ImageNet images from equivalent classes \cite{cinic}. For example, images from the {\em airplane, aeroplane, plane (airliner)} and {\em airplane, aeroplane, plane (bomber)} ImageNet classes were allocated to the {\em airplane} \texttt{CIFAR10} top-level class. Here we use only the 210{,}000 images taken from ImageNet.

{\bf \texttt{ImageNet-Far}:} Finally, as stronger exemplar of distributional shift, we built ImageNet-Far. As above,  we used rescaled ImageNet images, but chose classes that might not be under direct inheritance from the \texttt{CIFAR10}-synonymous classes. For example, for the \texttt{CIFAR10} label {\em deer}, we included the ImageNet categories {\em ibex}, {\em gazelle}, and for the \texttt{CIFAR10} label {\em horse} we included the ImageNet category {\em zebra}, which was not included in \texttt{CINIC10}.

\smallsec{Generalization Measures}
We evaluate each model on each test set in terms of both \textit{accuracy} and \textit{crossentropy}. Accuracy remains a centrally important measure of classification performance for the task of out-of-sample generalization. As accuracy ignores the probability assigned to a guess, we also employ the crossentropy metric to evaluate model behavior: how confident it is in its top prediction, and whether its distribution over alternative categories is sensible. Note that this interpretation arises naturally when computing crossentropy with a one-hot vector, as only the probability mass allocated to the ground-truth choice contributes to the score. Crossentropy becomes even more informative when computed with respect to human soft labels that distribute the mass unlike one-hot vectors. In this case, the second guess of the network, which provides a sense of the most confusable classes for an image, will likely be a large secondary contributor to the loss. To provide a more readily interpretable heuristic measure of this, we introduce a new accuracy measure called \textit{second-best accuracy} (SBA). While top-1 accuracy may largely asymptote, we expect that gains in SBA may still have a way to go.
\begin{figure*}[!t]
    \centering
    \includegraphics[width=1.0\linewidth]{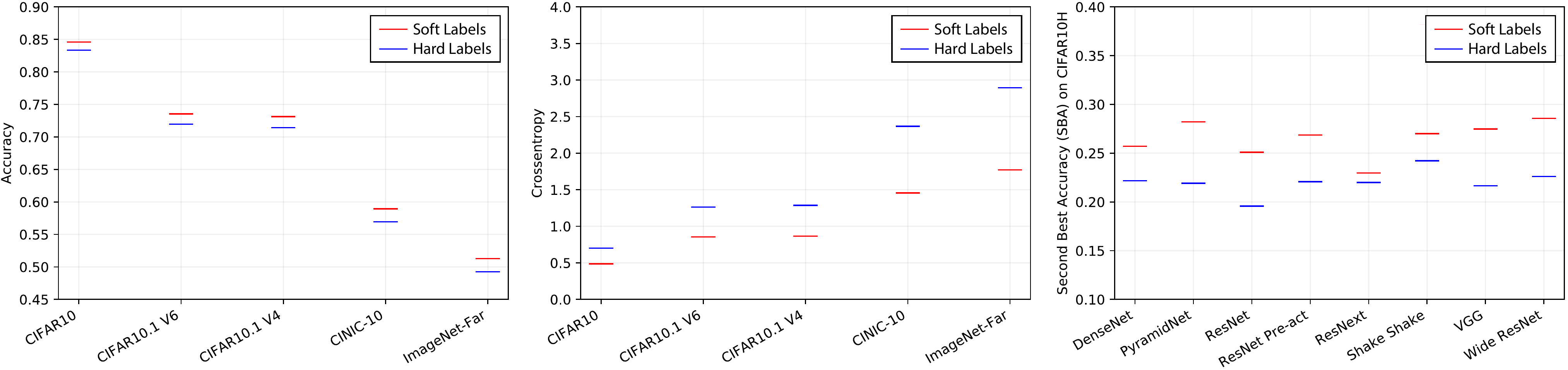}
    \caption{Generalization results. Left: accuracy against ground-truth labels, for increasingly out-of-training-sample distributions, averaged across CNNs. Accuracy was higher using human labels for every individual CNN and dataset. Center: crossentropy against ground-truth labels, averaged across CNNs. Loss was lower using human labels for every individual CNN and dataset. Right: Second best accuracy (SBA) for all models using \texttt{CIFAR10H} held out set, averaged across folds.}
    \label{fig:generalization_results}
\end{figure*}

\subsection{Human Labels Improve Generalization}
\label{acc_imp_results}

\begin{figure}[!t]
    \centering
    \includegraphics[width=1.0\linewidth]{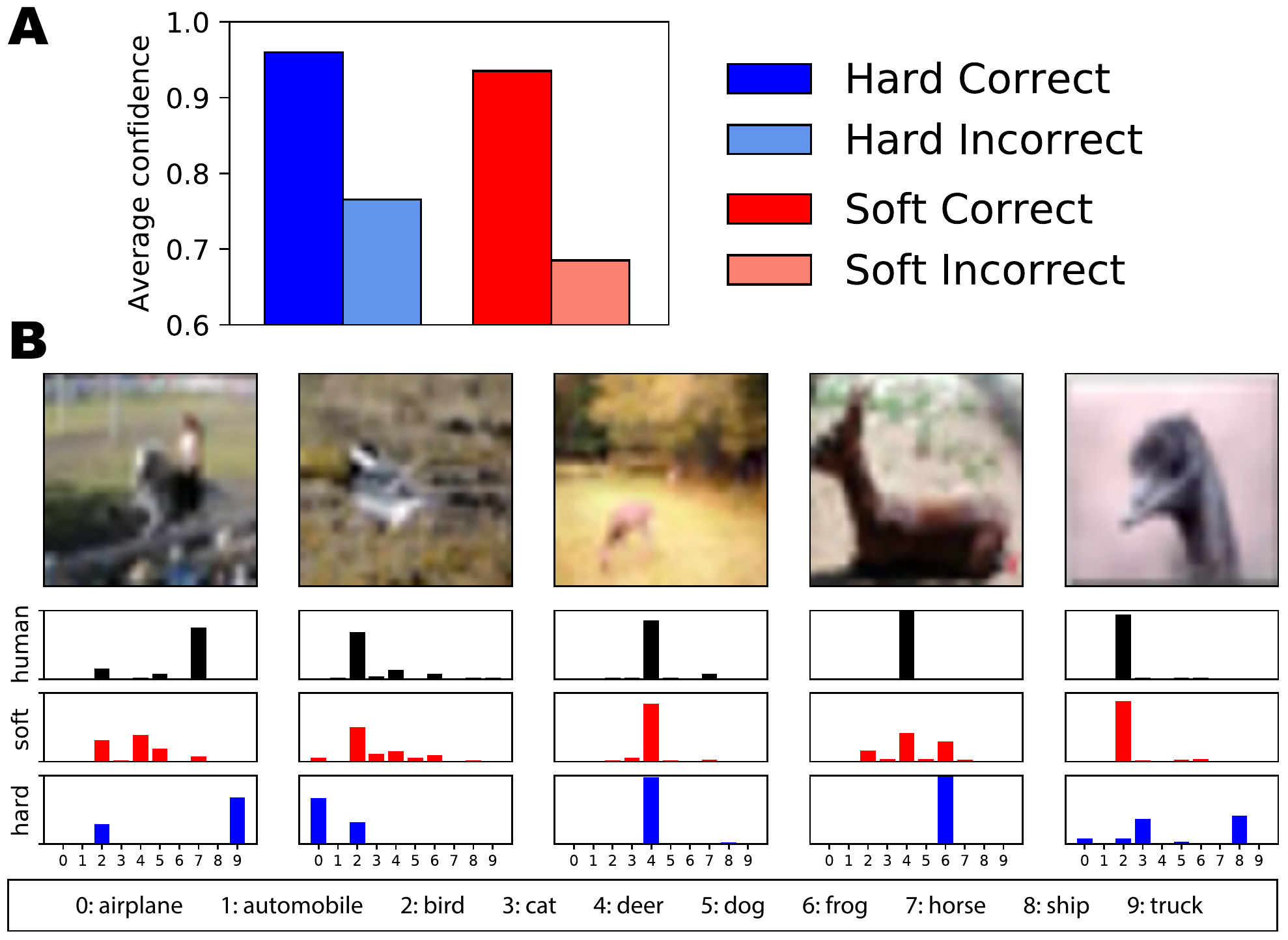}
    \caption{(A) Mean confidence for correctly/incorrectly classified examples after hard/soft label training. Soft-label models are far less confident when incorrect than hard-label controls, and only slightly less confident when correct. (B) Soft label training yields predictions that distribute probability mass more like people, with the same top choice.}
    \label{fig:predictions_detail}
    \vspace{-1mm}
\end{figure}

We train each CNN described above on both one-hot labels (default, control) and on \texttt{CIFAR10H} soft human labels (ours), and evaluate on each of the proposed test sets with increasingly out-of-sample distributions.

Our first finding is that when we train CNNs on \texttt{CIFAR10H} soft labels, their accuracy improves on all generalization datasets compared to our control (Figure \ref{fig:generalization_results}, left). This pattern was replicated across individual cross-validation folds for every individual model (not shown). A key feature of this boost in generalization is that it increases as test datasets become increasingly out-of-training-distribution (horizontal axis, left to right). For example, while using human soft labels gives us only a 1\% improvement (from 83.5\% to 84.5\%) when evaluated on \texttt{CIFAR10}, the same models when evaluated on ImageNet-Far achieved an accuracy gain of 2\% on average (from 49.4\% to 51.4\%).

This pattern is even more evident when we consider the crossentropy metric (Figure \ref{fig:generalization_results}, center). For example, while using human soft labels gives us a 29\% reduction in crossentropy (from 0.7 to 0.5) when evaluated on \texttt{CIFAR10}, the same models when evaluated on ImageNet-Far achieve a reduction of 38\% on average (from 2.9 to 1.8). These results imply that models trained on our soft labels show better confidence in their correct choices, and allocate more probability to the ground-truth during errors. 

Finally, CNNs trained on our soft labels consistently show significant boosts in SBA compared to controls, performing on average 5\% better (Figure \ref{fig:generalization_results}, right). This shows improvement in generalization in a broader sense: the distribution of the most likely two categories has important consequences for the graceful degradation in generalization we hope a good model provides, as well as for the nature of guesses made by a classification model when it is wrong.

Figure \ref{fig:predictions_detail} provides an additional picture of model behavior on our validation folds beyond overall generalization performance. Encouragingly, we find that soft-label-trained models are significantly less confident when incorrect than hard-label-trained controls, but only marginally less confident when correct (Figure \ref{fig:predictions_detail}a), and more generally provide a better fit to patterns of human uncertainty (Figure \ref{fig:predictions_detail}b).

\begin{table}[!h]
\scriptsize
\begin{center}

\begin{tabular}{lcccc}
\\
\hline
ResNet~\cite{he2016deep} & \tenh & \ten & \vfour & \vsix \\
\hline
Trained \cifar & 0.82 & 0.25 & 0.84 & 0.82 \\
FT \cifar & 0.57 & 0.19 & 0.60 & 0.58 \\
FT \cifar\phantom{.} with \textit{mixup}~\cite{zhang2017mixup} & 0.36 & {\bf 0.18} & {\bf 0.48} & 0.46 \\
FT \cifarh category soft targets & 0.42 & 0.21 & 0.53 & 0.51 \\
FT \cifarh soft targets (ours) & {\bf 0.35} & 0.19 & 0.50 & 0.49 \\
FT \cifarh sampled hard targets (ours) & {\bf 0.35} & 0.19 & {\bf 0.48} & {\bf 0.46} \\
\end{tabular}
\vspace{-2mm}

\begin{tabular}{lcccc}
\\
\hline
ResNet preact~\cite{he2016identity} & \tenh & \ten & \vfour & \vsix \\
\hline
Trained \cifar & 0.75 & 0.20 & 0.69 & 0.66 \\
FT \cifar & 0.65 & 0.19 & 0.61 & 0.59 \\
FT \cifar\phantom{.} with \textit{mixup}~\cite{zhang2017mixup} & 0.40 & {\bf 0.18} & 0.45 & 0.43 \\
FT \cifarh category soft targets & 0.44 & 0.23 & 0.47 & 0.46 \\
FT \cifarh soft targets (ours) & 0.35 & 0.21 & 0.49 & 0.48 \\
FT \cifarh sampled hard targets (ours) & {\bf 0.34} & 0.19 & {\bf 0.42} & {\bf 0.41} \\
\end{tabular}
\vspace{-2mm}

\begin{tabular}{lcccc}
\\
\hline
VGG~\cite{simonyan2014very} & \tenh & \ten & \vfour & \vsix \\
\hline
Trained \cifar & 0.71 & 0.26 & 0.79  & 0.76  \\
FT \cifar & 0.54  & {\bf 0.20}  & 0.62  & 0.59 \\
FT \cifar\phantom{.} with \textit{mixup}~\cite{zhang2017mixup} & 0.47  & {\bf 0.20} & 0.56  & 0.53 \\
FT \cifarh category soft targets & 0.42  & 0.22  & 0.51  & 0.49 \\
FT \cifarh soft targets (ours) & {\bf 0.34}  & 0.21  & {\bf 0.49}  & 0.48 \\
FT \cifarh sampled hard targets (ours) & 0.35  & 0.21 & {\bf 0.49}  & {\bf 0.47} \\
\end{tabular}
\vspace{-2mm}

\begin{tabular}{lcccc}
\\
\hline
DenseNet~\cite{huang2017densely} & \tenh & \ten & \vfour & \vsix \\
\hline
Trained \cifar & 0.61 & 0.15 & 0.54 & 0.54  \\
FT \cifar & 0.59 & 0.14 & 0.51 & 0.50 \\
FT \cifar\phantom{.} with \textit{mixup}~\cite{zhang2017mixup} & 0.36 & {\bf 0.13} & 0.43 & 0.42 \\
FT \cifarh category soft targets & 0.39 & 0.18 & 0.42 & 0.42 \\
FT \cifarh soft targets (ours) & 0.32 & 0.17 & {\bf 0.40} & 0.40 \\
FT \cifarh sampled hard targets (ours) & {\bf 0.31} & 0.16 & {\bf 0.40} & {\bf 0.39} \\
\end{tabular}
\vspace{-2mm}

\begin{tabular}{lcccc}
\\
\hline
PyramidNet~\cite{han2017deep} & \tenh & \ten & \vfour & \vsix \\
\hline
Trained \cifar & 0.54 & 0.12 & 0.42 & 0.42 \\
FT \cifar & 0.51 & {\bf 0.11} & 0.38 & 0.38 \\
FT \cifar\phantom{.} with \textit{mixup}~\cite{zhang2017mixup} & 0.49 & {\bf 0.11} & 0.40 & 0.40 \\
FT \cifarh category soft targets & 0.36 & 0.14 & {\bf 0.32} & {\bf 0.32} \\
FT \cifarh soft targets (ours) & {\bf 0.28} & 0.13 & 0.35 & 0.34 \\
FT \cifarh sampled hard targets (ours) & {\bf 0.28} & 0.12 & {\bf 0.32} & {\bf 0.32} \\
\end{tabular}
\vspace{-2mm}

\begin{tabular}{lcccc}
\\
\hline
ResNext~\cite{xie2016aggregated} & \tenh & \ten & \vfour & \vsix \\
\hline
Trained \cifar & 0.47 & {\bf 0.10} & 0.37 & 0.36  \\
FT \cifar & 0.46 & {\bf 0.10} & 0.35 & 0.34 \\
FT \cifar\phantom{.} with \textit{mixup}~\cite{zhang2017mixup} & 0.47 & {\bf 0.10} & 0.37 & 0.36 \\
FT \cifarh category soft targets & 0.37 & 0.17 & 0.37 & 0.36 \\
FT \cifarh soft targets (ours) & 0.29 & 0.13 & {\bf 0.34} & {\bf 0.33} \\
FT \cifarh sampled hard targets (ours) & {\bf 0.28} & 0.13 & {\bf 0.34} & {\bf 0.33} \\
\end{tabular}
\vspace{-2mm}

\begin{tabular}{lcccc}
\\
\hline
Wide ResNet \cite{zagoruyko2016wide} & \tenh & \ten & \vfour & \vsix \\
\hline
Trained \cifar & 0.46 & 0.14 & 0.40 & 0.39 \\
FT \cifar & 0.42 & {\bf 0.12} & 0.37 & 0.36 \\
FT \cifar\phantom{.} with \textit{mixup}~\cite{zhang2017mixup} & 0.40 & {\bf 0.12} & 0.37 & 0.36 \\
FT \cifarh category soft targets & 0.36 & 0.15 & 0.33 & 0.33 \\
FT \cifarh soft targets (ours) & {\bf 0.27} & 0.13 & 0.32 & 0.31 \\
FT \cifarh sampled hard targets (ours) & 0.28 & 0.13 & {\bf 0.31} & {\bf 0.30} \\
\end{tabular}
\vspace{-2mm}

\begin{tabular}{lcccc}
\\
\hline
Shake-Shake~\cite{shake-shake} & \tenh & \ten & \vfour & \vsix \\
\hline
Trained \cifar & 0.60 & 0.09 & 0.34 & 0.33 \\
FT \cifar & 0.51 & {\bf 0.07} & 0.28 & 0.27 \\
FT \cifar\phantom{.} with \textit{mixup}~\cite{zhang2017mixup} & 0.63 & 0.08 & 0.34 & 0.33 \\
FT \cifarh category soft targets & 0.33 & 0.12 & 0.28 & 0.28 \\
FT \cifarh soft targets (ours) & {\bf 0.26} & 0.10 & {\bf 0.27} & {\bf 0.26} \\
FT \cifarh sampled hard targets (ours) & 0.27 & 0.10 & {\bf 0.27} & 0.27 \\
\end{tabular}

\end{center}
\caption{Crossentropy for each holdout set (columns from left to right: holdout human soft labels (\texttt{c10H}), holdout ground truth labels (\texttt{c10}), the entire \texttt{CIFAR10.1v4} dataset, and the entire \texttt{CIFAR10.1v6} dataset. Crossentropy for our human labels decreases substantially after fine-tuning (FT), especially when using human targets. Fine-tuning on human targets also produces the best generalization in terms crossentropy on \texttt{CIFAR10.1}.}
\label{benchmark-table}

\end{table}

\section{Alternative Soft Label Methods}
\label{sec:alternatives}
Above, we show out-of-sample classification benefits arise from training on our human labels.
One natural question that arises is whether this improvement is the result of simply training with soft labels (\textit{i.e.,} allowing the model to distribute the probability mass over more than one class), or due to the fact that this distribution explicitly mimics human uncertainty. Here we show the answer is the latter.

\subsection{Setup}
\smallsec{Training}
We set out to demonstrate that training with human labels provides benefits even over competitive baselines. We use the same CNN architectures and setup as in Section~\ref{model-steup} with one notable exception: we pre-train the networks before incorporating the soft labels (this allows us to achieve the best possible fit to humans). To do so, we train using the standard \texttt{CIFAR10} training protocol using 50{,}000 images and the optimal hyperparameters in the repository, either largely replicating or surpassing the original accuracies proposed in the papers for each architecture. We then fine-tune each pretrained model using either hard-label controls or our human soft labels on the \texttt{CIFAR10} test set. This fine-tuning phase mirrors the training phrase from Section~\ref{model-steup}: we used 10-folds, trained for 150 epochs, and searched over learning rates $0.1$, $0.01$, and $0.001$.

\vspace{-2mm}
\smallsec{Evaluation} We evaluate the results on the holdout folds of \texttt{CIFAR10H} with both human soft labels and ground truth hard labels, as well as on the ground truth hard labels of both the \texttt{CIFAR10.1v4} and \texttt{CIFAR10.1v6} datasets. We also  shift our attention to evaluating crossentropy rather than accuracy. With \texttt{CIFAR10} pretraining, the accuracy of all models is high, but this gives no indication of the level of confidence or the ``reasonableness'' of  errors. Crossentropy, on the other hand, does exactly that: measures the level of confidence when evaluated on hard labels and the ``reasonableness'' of errors when evaluated on human soft labels.

\subsection{Methods}
To test for simpler and potentially equally effective alternatives to approximating the uncertainty in human judgments, we include a number of competitive baselines below. 

\vspace{-2mm}
\smallsec{Ground Truth Control}
The first baseline we consider is a ``control'' fine-tuning condition where we use identical image data splits, but fine-tune using the ground-truth hard labels. This is expected to improve upon the pretrained model as it utilizes the additional $9{,}000$ images previously unseen.

\vspace{-2mm}
\smallsec{Class-level Penalty}
One much simpler alternative to image-level human soft labels is class-level soft labels. That is, instead of specifying how much each image resembles each category, we could simply specify which classes are more confusable on average using a class-level penalty. However, while we know, for example, that dogs and cats are likely more confusable on average than dogs and cars, it's not clear what the optimal class-level penalties should be. Since exhaustively searching for competitive inter-class penalties is inefficient, we propose to generate gold-standard penalties by summing and re-normalizing our human probabilities within each class (\textit{i.e.,} resulting in exactly 10 unique soft-label vectors). This also allows us to determine if image-level information in our human soft labels is actually being utilized as opposed to class-level statistics across image exemplars. In this baseline, fine-tuning simply uses these greatly compressed soft vectors as targets.

\vspace{-2mm}
\smallsec{Knowledge Distillation} 
As discussed in Section \ref{section:related_work}, softmax probabilities of a trained neural network can be used as soft labels because they contain information inferred by the network about the similarity between categories and among images. The pretrained networks from this section provide such probabilities and so provide a corresponding baseline. However, we can infer from the results in Section \ref{acc_imp_results} that hard-label-trained CNNs infer class probabilities that do not approximate those of humans, because incorporating explicit supervision to humans provides different results in terms of generalization. So, to provide a stronger baseline in this respect, we include an ensemble of the predictions from all eight models (\textit{i.e.,} providing soft predictions due to uncertainty from variation across models).

\vspace{-2mm}
\smallsec{\textit{mixup}}
\textit{mixup} is a technique for soft label generation that improves the generalization of natural image classification models trained on \texttt{CIFAR10} among others \cite{zhang2017mixup}---see Section \ref{section:related_work}. As such, it provides an interesting and competitive baseline with which to compare training with human soft labels. Concretely, \textit{mixup} generates soft labels by taking convex combinations of pairs of examples, encouraging linear behavior between them. These combinations constitute virtual training examples ($\bar{x}$, $\bar{y}$) that are sampled from a vicinial distribution, and take on the form
\begin{align*}
    \bar{x} &= \lambda x_i + (1-\lambda)x_j\\
    \bar{y} &= \lambda y_i + (1-\lambda)y_j,
\end{align*}
where ($x_i$, $x_j$) are examples from the dataset, and ($y_i$, $y_j$) are their labels. The strength of the interpolation $\lambda \in [0,1]$ is sampled according to $\text{Beta}(\alpha, \alpha)$, where $\alpha$ is a hyperparameter. For our \textit{mixup} baseline, we apply this procedure to the ground truth labels corresponding to each of the same 10 splits used above. For each architecture, we searched for the best value of $\alpha$ from $0.1$ to $1.0$ in increments of $0.1$.

\vspace{-2mm}
\smallsec{Soft Labels Versus Sampling}
Finally, we run one additional experiment beyond the soft label baselines above. Results from Section \ref{section:primary_training} suggest that human soft labels are useful, but how should we best incorporate them into training? In Section \ref{section:theory}, we justified using human probabilities as targets to minimize the expected loss. However, another valid option is to sample from $p_{\rm hum}(y|x)$, \textit{i.e.,} sample one-hot labels from categorical distribution parameterized by the human probabilities conditioned on each image. If we sample a new label each time the image is presented to the network for a new gradient update, the label uncertainty will still be incorporated, but there will be additional variation in the gradients that could act as further regularization. To test for any such advantages of label sampling, we fine-tuned a second corresponding set of models using this method, sampling a new label for each image on each epoch.

\subsection{Human Soft Labels Beat Alternatives}

Results are summarized for each architecture and method in Table \ref{benchmark-table}. The first column is our primary measure of fit to humans; the last two assess further generalization.

Note that for pretrained models (first row of each sub-table) crossentropy to ground truth labels is always lower than human soft labels, verifying what we expected: human soft labels provide additional information that is not inferred via training with ground truth. This is a first test that the information (informative probabilities) usually inferred by these networks using hard labels (\textit{i.e.,} knowledge distillation) does not agree with humans. We further tested an ensemble of all eight networks in the top rows (\textit{i.e.,} with no fine-tuning on human soft labels), and while this model is more like humans than any individual hard-label-trained model (crossentropy is 0.41), it is still not a substitute for human supervision. The benefit from our labels also appears to manifest during generalization, as in the last two columns (\textit{i.e.,} \texttt{v4} and \texttt{v6} holdout sets) they show higher crossentropy than alternative approaches. Next, looking at the same top rows, note that there is little correspondence between recency of the architecture and fit to humans. In fact, Shake-Shake is the state-of-the-art of the eight yet is not one of the top three models in terms of fit to humans.

In the remaining rows of each sub-table, we can see an increase in fit to humans using our various fine-tuning schemes. This is expected in all cases given that all of these models are ultimately given more data than pretrained models. However, not all fine-tuning methods are equally effective. Importantly, fit to humans (second column) is best when either using our image-level soft labels or sampling hard labels using them (bottom two rows). Interestingly, category soft labels (4th rows) were also effective, but to a lesser degree. \textit{mixup} was more effective than using ground truth labels alone, but less effective than any methods using human information. Lastly, we note that, while omitted for brevity, we found no loss in accuracy when using human labels in any of the conditions that utilized them.

\begin{table}
    \begin{center}
        \begin{tabular}{l c c c c}
        & \multicolumn{2}{c}{Accuracy} & \multicolumn{2}{c}{Crossentropy} \\
        Architecture & \texttt{C10} & \texttt{C10H} & \texttt{C10} & \texttt{C10H} \\
        \hline  \\ [-2.1ex]
        VGG & 7\% & {\bf 8\%} & 7.9 & {\bf 4.1} \\
        DenseNet  & 17\% & {\bf 19\%} & 6.9 & {\bf 3.0} \\ 
        PyramidNet  & {\bf 22\%} & 19\% & 5.7 & {\bf 2.8} \\
        ResNet  & 15\% & {\bf 23\%} & 6.1 & {\bf 3.1} \\
        ResNext  & {\bf 25\%} & 24\% & 4.2 & {\bf 2.7} \\
        Wide ResNet  & 24\% & {\bf35\%} & 4.1 & {\bf 2.2} \\
        ResNet preact  & 17\% & {\bf29\%} & 6.3 & {\bf 2.6} \\
        Shake-Shake  & {\bf 39\%} & {\bf 39\%} & 4.0 & {\bf 2.1}
        \end{tabular}
    \end{center}
    
    \caption{Accuracy and crossentropy after FGSM attacks on the \texttt{CIFAR10}-tuned (baseline) and \texttt{CIFAR10H}-tuned networks. Using human labels always results in lower (better) crossentropy, and in the majority of cases, higher accuracy.}
    \label{fgsm-table}
    \vspace{-2mm}
\end{table}

\section{Robustness to Adversarial Attacks}
Because our soft labels contain information about the similarity structure of images that is relevant to the structure of perceptual boundaries, we might expect that representations learned in service of predicting them would be more robust to adversarial attacks, particularly in cases where similar categories make for good attack targets. Moreover, subsequent explorations of knowledge distillation \cite{Hinton2015DistillingTK,papernot2016distillation} have demonstrated that such practices can support adversarial robustness. If human judgments of perceptual similarity are superior to those inferred by CNNs---in the form of $p(y|x)$---we would expect distillation of human knowledge into a CNN would at the very least also increase robustness.

\vspace{-2mm}
\smallsec{Setup}
We use the same pretrained and fine-tuned (hard versus soft) models from Section~\ref{sec:alternatives}. To measure robustness after each training scheme, we evaluate both accuracy and crossentropy (the latter again being a more sensitive measure of both confidence and entropy) against the hard class labels. As attack methods, we evaluate two additive noise attacks: the Fast Gradient Sign Method (FGSM) \cite{kurakin2016adversarial}, and Projected Gradient Descent (PGD) \cite{pgd_kurakin2016adversarial}, using the \texttt{mister\_ed} toolkit\footnote{\url{github.com/revbucket/mister_ed/}} for \texttt{PyTorch}. For both methods, we explored $\ell_\infty$ bounds of $4$ to $8$ in increments of $1$. Since we found no significant differences in the results, we report all attack results using a constant $\ell_\infty$ bound of $4$ for brevity. 

\vspace{-2mm}
\smallsec{Human Soft Labels Confer Robustness}
FGSM results are reported in Table \ref{fgsm-table}, averaged over all $10{,}000$ images in the \texttt{CIFAR10} test set. In all cases, crossentropy (which attack methods seek to maximize) is much lower (roughly half) after attacking the human-tuned network compared to fine-tuning with original one-hot labels. For five out of eight architectures, accuracy also improves when using human soft targets. The two largest differences (Wide Resnet and ResNet preact) favor the human labels as well. Note that no explicit (defensive) training was required to obtain these improvements beyond previous training with human labels.

\begin{figure}[!t]
    \centering
    \includegraphics[width=0.65\linewidth]{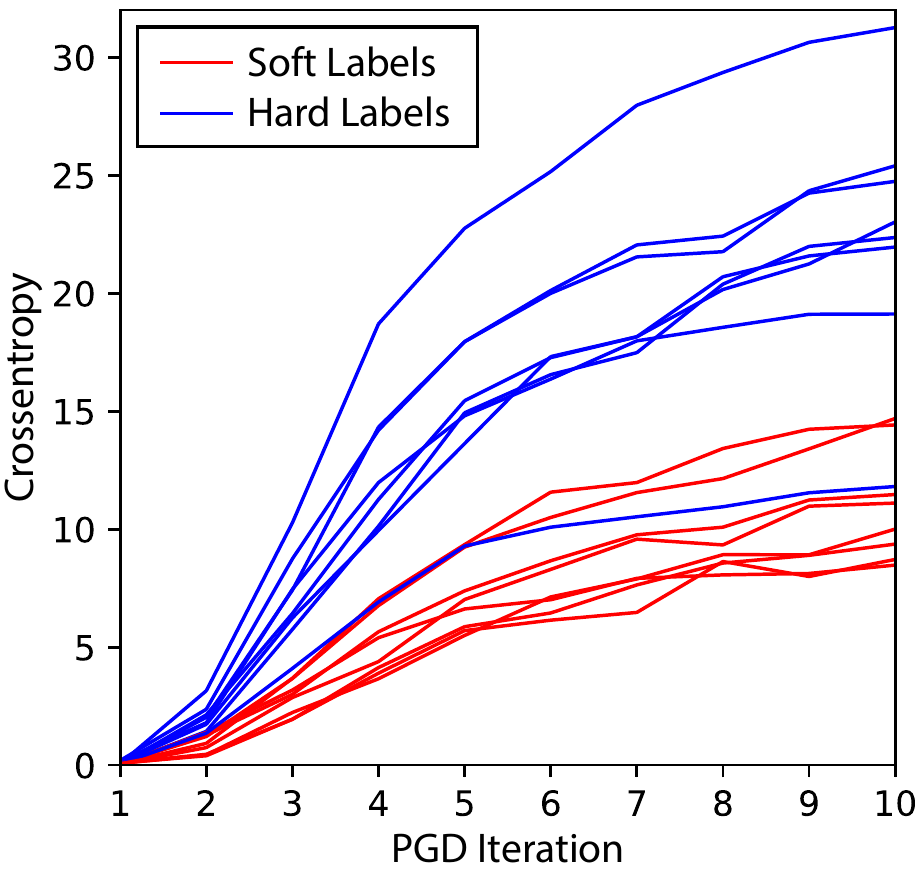}
    \caption{Crossentropy as a function of PGD iteration. Successive iterations increase crossentropy as expected, but more slowly after soft-label fine-tuning.}
    \label{fig:pgd_curves}
    \vspace{-2mm}
\end{figure}

Without active defensive training, PGD is expected to drive accuracy to $0\%$ given enough iterations. To explore the intrinsic defenses of our two label-training conditions to PGD attacks, we plot the increase in loss for each architecture and label-training scheme in Figure \ref{fig:pgd_curves}. While accuracy was driven to $0\%$ for each network when trained on standard labels, and $1\%$ for each network with human labels, loss for the former is driven up much more rapidly, whereas the latter asymptotes quickly. Put simply, a much higher degree of effort is required to successfully attack networks that behave more like humans.
\section{Discussion}
In this work, we have demonstrated that incorporating information about human category uncertainty at the image-level can help protect against the perils of distributional shift and adversarial attacks. Notably, common classification benchmarks often do not naturally provide such protections on their own \cite{torralba2011unbiased}. Further, besides explicitly incorporating this information, it gives a way of measuring whether our learning algorithms are inferring good similarity structure (beyond just top-1 performance). If we can begin to find good learning procedures that derive such information, we can obtain human-like robustness in our models without the need of explicit human supervision. However, developing such a robust models will take significant time and research---our dataset provides a first step (an initial gold standard with respect to a popular benchmark) in measuring this progress, even when not used for training.

Although our data collection method does not immediately seem to scale to larger training sets, it's certainly possible to collect informative label distributions at a cost comparable to what we often spend on compute to find better top-1-fitting architectures. Interestingly, we found that the bulk of human uncertainty is concentrated in approximately 30\% of the images in our dataset, meaning straightforward and much more efficient methods for mining only these more informative labels can be employed. In any case, we see the main contribution of such datasets as testing environments for algorithms intended for much larger datasets.

\vspace{-2mm}
\smallsec{Acknowledgements} This work was supported by grant number 1718550 from the National Science Foundation.

\clearpage
{\small
\bibliographystyle{ieee_fullname}
\bibliography{references}
}

\end{document}